# DeepReduce: A Sparse-tensor Communication Framework for Distributed Deep Learning

Kelly Kostopoulou
Columbia University
New York, NY, USA
kelkost@cs.columbia.edu

Hang Xu
KAUST
Saudi Arabia
hang.xu@kaust.edu.sa

Aritra Dutta
KAUST
Saudi Arabia
aritra.dutta@kaust.edu.sa

Xin Li
University of Central Florida
Orlando. FL, USA
xin.li@ucf.edu

Alexandros Ntoulas
National and Kapodistrian
University of Athens, Greece
antoulas@di.uoa.gr

Panos Kalnis
KAUST
Saudi Arabia
panos.kalnis@kaust.edu.sa

## Abstract

Sparse tensors appear frequently in distributed deep learning, either as a direct artifact of the deep neural network's gradients, or as a result of an explicit sparsification process. Existing communication primitives are agnostic to the peculiarities of deep learning; consequently, they impose unnecessary communication overhead. This paper introduces DeepReduce, a versatile framework for the compressed communication of sparse tensors, tailored for distributed deep learning. DeepReduce decomposes sparse tensors in two sets, values and indices, and allows both independent and combined compression of these sets. We support a variety of common compressors, such as Deflate for values, or run-length encoding for indices. We also propose two novel compression schemes that achieve superior results: curve fitting-based for values and bloom filter-based for indices. DeepReduce is orthogonal to existing gradient sparsifiers and can be applied in conjunction with them, transparently to the end-user, to significantly lower the communication overhead. As proof of concept, we implement our approach on Tensorflow and PyTorch. Our experiments with large real models demonstrate that DeepReduce transmits fewer data and imposes lower computational overhead than existing methods, without affecting the training accuracy.

## 1 Introduction

Training complex deep neural networks (DNNs) with large datasets, requires distributed computing infrastructure. A cluster of powerful compute nodes, each equipped with one or multiple accelerators (e.g., GPUs), independently process partitions of the data, and synchronize regularly by exchanging the model parameters or the gradients. Data exchange results in network bottleneck [51, 55, 59] that becomes more prominent for larger DNNs (e.g., GPT-3 [13] has $3 \cdot 10^8$ to $175 \cdot 10^9$ parameters), and for higher ratios of computation to communication speed (e.g., faster GPUs such as the NVidia A100).

Lossy compression is commonly employed to alleviate the network bottleneck. Intuitively, the stochastic nature of the training optimization algorithm (e.g., some variant of Stochastic Gradient Descend [58]), tolerates the small information loss introduced by lossy compression. Three[1] main families of compression methods exist (refer to [75] for a survey): (i) Quantization [7, 11, 20, 44, 61? ], where each tensor element is replaced by a lower precision one (e.g., float8 instead of float32), to achieve in practice compression ratios in the order of 4x to 8x [75]. (ii) Sparsification [5, 49, 63, 65, 68, 72], where only a few elements (e.g., top-$r$ or random-$r$) of the tensor are selected; it can achieve compression ratios of 100x or more. (iii) Hybrid methods [10, 37, 48, 64], which combine quantization with sparsification to achieve higher compression ratios.

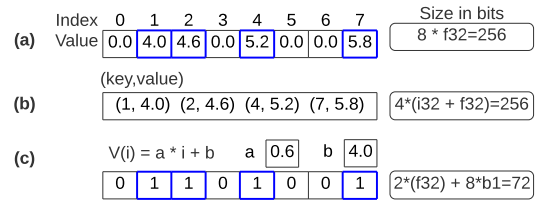

Figure 1: (a) Dense tensor format needs 256 bits; (b) sparse $\langle key, value \rangle$ form also needs 256 bits; (c) our method sends parameters $a$ and $b$ as well as an 8-bit string, i.e., 72 bits in total.

Consequently, the data exchanged during the training of DNN often correspond to *sparse* tensors, that is, tensors with many zero-value elements. Sparse tensors may: (i) be generated explicitly by sparsification; or (ii) be direct artifacts of the training process; for instance, the gradients of the NCF model [32] consist of roughly 40% zero elements (Section 6). Figure 1.a depicts an example tensor containing 8 real values, 4 of which are zero. Assuming that each real value is represented by 32 bits, the dense representation of the array requires $8 \times 32 = 256$ bits. In an attempt to take advantage of sparsity, let us represent the array as a set of $\langle key, value \rangle$ pairs (see Figure 1.b), where $key$ is the index. Observe that, if indices are 32-bit integers, the *sparse* representation also needs $4 \times (32 + 32) = 256$ bits, negating the benefit of sparsification. Can we do better? In Figure 1.c we consider the indices as an ordered list represented by a boolean array of 8 bits, such that the $i^{th}$ bit is '1' if and only if the corresponding gradient element is non-zero. Moreover, we fit a function $V(i) = a \cdot i + b$ to the gradient values, obtaining parameter values $a = 0.6$ and $b = 4.0$. By transmitting only the bit string and parameters $a$, $b$, we are able to perfectly reconstruct the original tensor, while requiring only $8 \times 1 + 2 \times 32 = 72$ bits.

---
[1] There also exist methods that decompose the tensor into low-rank components [16, 70, 71], but are not common in practice due to the high computational cost.

The example demonstrates the significant margin for additional compression for sparse tensors. Recent works (e.g., SKCompress [39]) take advantage of these opportunities, but rely on a tightly coupled index and value compression algorithm that benefits only some scenarios (see Section 6). In practice there are complex trade-offs among data volume, model accuracy and computational overhead, in conjunction with system aspects, such as the network bandwidth and the communication library (e.g., NCCL [2] or Gloo [30]). Given that no solution fits all scenarios, practitioners need flexibility to adjust the way sparse tensors are compressed and transmitted, for each particular DNN model and system architecture.

In this paper, we propose DeepReduce, a framework for transmitting large sparse tensors through the network, tailored for distributed DNN training. DeepReduce decomposes the sparse tensor into two parts, indices and values, and allows for both independent and combined compression. By decoupling indices from values, it enables the effortless combination of a variety of compressors in a way that benefits each particular DNN and training setup. DeepReduce resides between the machine learning framework (e.g., Tensorflow, PyTorch) and the communication library. It exposes an easy-to-use API that encapsulates a variety of existing methods, such as Run Length [74] and Huffman encoding [36] for index compression; as well as Deflate [21] and QSGD [7] for value compression. DeepReduce also provides an index reordering abstraction, which is useful for combining value with index compressors. Moreover, we develop two novel compressors for sparse tensors: (i) a Bloom filter-based index compressor that reduces the size of keys by 50%, compared to the traditional $\langle key, value \rangle$ sparse representation; and (ii) a value compressor based on polynomial or double exponential curve fitting that reduces the values array to just a few parameters. Both of our compressors do not affect the quality of the trained model.

Our contributions include: (i) Our DeepReduce framework, described in Section 3; our code is available at https://github.com/hangxu0304/DeepReduce. (ii) Our novel index and value compressors, introduced in Sections 4 and 5, respectively. (iii) A comprehensive evaluation on a variety of large-scale models (Section 6) focused on practical applicability issues.

## 2 Background

**Notations.** By $[d]$ we denote the set of $d$ natural numbers $\{1, 2, \cdots, d\}$. To denote the cardinality and complement of a set $X$, we use $|X|$ and $X^c$, respectively. $x[i]$ denotes the $i^{\text{th}}$ component of vector $x$. By $x_I \in \mathbb{R}^d$ we denote a $|I|$-sparse vector, where $I \subseteq [d]$ is its support. $\|x\|$ and $\|x\|_\infty$ denote the $\ell_2$ and $\ell_\infty$ norm of a vector $x$, respectively.

**Compressed distributed training via SGD.** In DNN training (back-propagation) with $n$ workers (i.e., compute nodes), to update the model parameter $x$, each worker $i$, at each iteration, calculates a stochastic gradient $g_t^i$ by processing an independent batch of data. Often, for efficient communication, the gradient is compressed to $\tilde{g}_t^i$ and is communicated to all workers, either through a parameter server [56], or through a peer-to-peer collective, like Allreduce [62]. The aggregated gradient $\tilde{g}_t = \frac{1}{n} \sum_{i=1}^n \tilde{g}_t^i$ is then transmitted to all workers, who update the parameters of their local model via: $x_{t+1} = x_t - \eta_t \tilde{g}_t$, where $\eta_t$ is the learning rate; the process repeats until convergence. The compressed gradient $\tilde{g}_t^i$ is generated by an random operator $C(\cdot) : \mathbb{R}^d \to \mathbb{R}^d$, called *Compressor* [63], that satisfies $\mathbb{E}_C \|x - C(x)\|^2 \leq \Omega \|x\|^2$, where $\Omega > 0$ is the compression factor and the expectation is taken over the randomness of $C$. If $\Omega = 1 - \delta$ and $\delta \in (0, 1]$, $C$ is a $\delta$-*compressor*, denoted by $C_\delta$.

**Sparsification.** A rank-1 tensor that has mostly zero components is said to be sparse. Many modern DNNs are inherently sparse [19, 41] and so are their gradients. Sparse gradients can also be generated by a compressor that sparsifies [49, 65, 68, 72] the gradient $g$ to generate $\tilde{g} \in \mathbb{R}^d$ via: $\tilde{g}[i] = C(g[i]) = g[i]$, if $\in S$, otherwise $\tilde{g}[i] = 0$. We assume that $S \subset [d]$ is the support set of $\tilde{g}$ such that $\tilde{g}$ is $r$-sparse if and only if $|S| = r$. Two common sparsifiers are Random-$r$ [63] and Top-$r$ [5, 8]; both are $\delta$-compressors.

**Run Length Encoding (RLE)** [74] is a *lossless* compressor in which consecutive occurrences of symbols are encoded as $\langle frequency, symbol \rangle$ tuples. For example, string "aaaabaa" is encoded as: (4, "a"), (1, "b"), (2, "a"). RLE is used to compress large sequences of repetitive data. In this work, we employ bit-level RLE, where symbols are 0 or 1, for index compression.

**Huffman encoding** [36] is a *lossless* scheme that assigns the optimal average decode-length prefix codes, using a greedy algorithm to construct a Huffman code tree. Higher frequency symbols are encoding with fewer bits. For instance, string "aaaabaacaabaa" generates mapping ("a", "b", "c") → (0, 10, 11) resulting to the following encoding: 0000**10**0011**00**1000. Huffman encoding has been used to compress DNN weights [29, 31], as well as sparse gradient indices (e.g., SKCompress [39]).

**Bloom Filter** [12] is a probabilistic data structure that represents the elements of a set $S$. Initially, it is a bit string $\mathcal{B}$ of $m$ bits, all set to 0. To insert an element $y_j \in S$ in the Bloom filter, we apply $k$ independent hash functions $h_i$ on $y_j$. Each $h_i$ yields a bit location in $\mathcal{B}$, and changes that bit to 1. To query if $y_j \in S$, we apply all $k$ hash functions on it. If $h_i(y_j) = 0$ for any $i \in [k]$, then the element *surely* does not belong to $S$ (i.e., no false negatives). In contrast, if $h_i(y_j) = 1$ for all $i \in [k]$, then $y_j$ may or may not belong to $S$ (i.e., possible *false positives*). The false positive rate (FPR) [12] of $\mathcal{B}$ is: $\epsilon \approx (1 - e^{-k|S|/m})^k$. Figure 2 illustrates a Bloom filter $\mathcal{B}$ with $m = 12$ bits and $k = 3$ hash functions, representing a set $S = \{y_1, y_2, y_3\}$. During querying, $y_1$ and $y_4$ are correctly identified as belonging (i.e., true positive) and not belonging (i.e., true negative) to $\mathcal{B}$, respectively. In contrast, $y_5$ is wrongly identified (i.e., false positive) as belonging to $\mathcal{B}$.

## 3 System Architecture

DeepReduce is optimized for data parallel distributed DNN training. It resides between the machine learning framework (e.g., TensorFlow, PyTorch) and the communication library, and is oblivious to the communication topology (e.g., parameter server, or Allreduce peer-to-peer collective). DeepReduce offers a simple API whose functions can be overridden to implement, with minimal effort, a wide variety of index and value compression methods for sparse tensors. The system architecture is shown in Figure 3. At the transmitting worker side, the input to DeepReduce is a sparse tensor, which is fed directly from the ML framework, for the case of inherently sparse models; or, is generated by an explicit sparsification process. In our implementation, we employ the GRACE [75] library for the sparsification operation, since it includes many popular sparsifiers; other libraries can also be used.

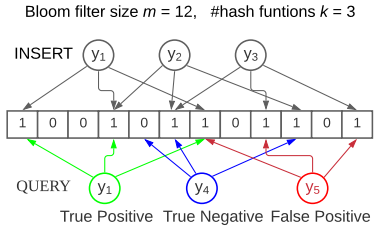
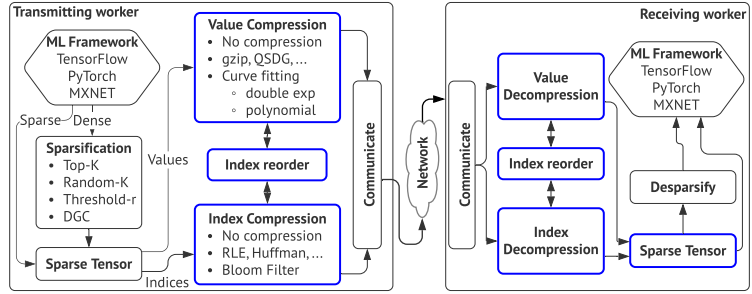

Figure 2: Bloom filter example.    Figure 3: System architecture. The DeepReduce framework is highlighted in blue.

Sparse tensors are typically represented as $\langle key, value \rangle$ tuples. DeepReduce decouples the keys from the values and constructs two separate data structures. Let $\tilde{g} \in \mathbb{R}^d$ be the sparse gradient, where $d$ is the number of model parameters and $\|\tilde{g}\|_0 = r$ is the number of nonzero gradient elements. Let $S$ be the set of $r$ indices corresponding to those elements. DeepReduce implements two equivalent representations of $S$: ($i$) an array of $r$ integers; and ($ii$) a bit string $B$ with $d$ bits, where $\forall i \in [1, d], B[i] = 1$ if and only if $\tilde{g}[i] \neq 0$. The two representations are useful for supporting a variety of index compressors; for example, the bit string representation is used by RLE [15]. The Index Compression module, depicted in Figure 3, encapsulates the two representations and implements several algorithms for index compression, including our Bloom filter based proposal, the existing RLE and Huffman encoders and an option to bypass index compression. Note that DeepReduce supports both lossless (e.g., RLE) and lossy (e.g., Bloom filter-based) compressors.

The Value Compression module receives the sparse gradient values and compresses them independently. Several compressors, such as Deflate [21] and QSGD [7], are implemented, in addition to our own polynomial and exponential curve fitting-based methods. Again, there is an option to bypass value compression. Some value compressors (e.g., our own proposals), require reordering of the gradient elements, which is handled by the Index reorder module. DeepReduce then combines in one container the compressed index and value structures, the reordering information and any required metadata; the container is passed to the communication library.

The receiving worker, at the right of Figure 3, mirrors the structure of the transmitter, but implements the reverse functions, that is, index and value decompression, and index reordering. The reconstructed sparse gradient is routed to GRACE for de-sparsification, or is passed directly to the ML framework. It is worth mentioning that DeepReduce is general enough to represent popular existing methods that employ proprietary combined value and index compression. For example, SKCompress [39] can be implemented in DeepReduce as follows: SketchML plus Huffman for values, no index reordering, and delta encoding plus Huffman for indices.

## 4 Bloom Filter for Indices

This section introduces our novel Bloom filter-based, lossy index compressor. Recall that $S$ is the set of $r$ indices (i.e., $|S| = r$) corresponding to the nonzero components of sparse gradient $\tilde{g}$. Let us insert each item of $S$ into a Bloom filter $\mathcal{B}$ of size $m$, using $k$ hash functions, as shown in Figure 2.

**Naïve Bloom filter.** Let $V$ be an array of size $r$ containing the elements of $\tilde{g}$ that are indexed by $S$; formally: $\forall i \in [1, r]$ : $V(i) = \tilde{g}[S[i]]$. DeepReduce transmits $V$ and $\mathcal{B}$. The receiver initializes $ptr = 1$ and reconstructs the gradient as follows:

```
1  for i = 1 to d do  /* all d elements of gradient g̃ ∈ ℝ^d */
2      if i ∈ B then  g̃[i] = V[ptr]; ptr++; else g̃[i] = 0
```

If $\mathcal{B}$ were *lossless*, the above mentioned algorithm would have perfectly reconstructed the gradient. However, in practice Bloom filters exhibit false positive (*FP*) responses. Assume a single FP at $ptr = j$; then, for $j \leq ptr \leq r$, every $V[ptr]$ value will be assigned to the wrong sparse gradient element. Therefore, FPs cause a disproportionately large error to the reconstructed sparse gradient, significantly affecting the quality of the trained model, as we show in Section 6.1.

**No-error approach: Policy P0.** To address the drawback of the naïve approach, let us initialize set $P = \emptyset$ and slightly modify the previous reconstruction algorithm:

```
1  for i = 1 to d do  /* all d elements of gradient g̃ ∈ ℝ^d */
2      if i ∈ B then  insert i in P
```

Now $P$ contains the union of the true and false positive responses of $\mathcal{B}$. The transmitting worker can execute this algorithm and determine $P$ prior to any communication. With that information, it constructs array $V$ as follows: $\forall i \in [1, |P|]$ : $V(i) = \tilde{g}[P[i]]$. Essentially, $V$ contains the gradient elements that correspond to both true and false positives. Therefore, the receiving worker can reconstruct *perfectly* the sparse gradient. The trade-off is increased data volume, since the size of $V$ grows to $|P| \geq r$. In our technical report [45] we show that $|P|$ is at most $\lceil r + (\frac{1}{2})^{-\frac{\log(\epsilon)}{\log(2)}} (d - r) \rceil$, where $\epsilon$ is the false positive rate, and approaches to $r$ as $\epsilon \to 0$.

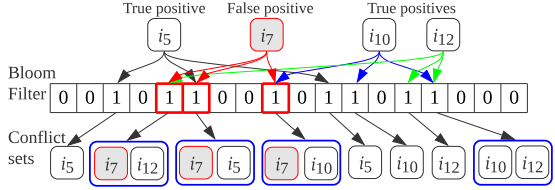

**Figure 4: Policy P2, with 3 hash functions and 8 conflict sets.**

The GRACE [75] sparsification library allows to use the original dense gradient $g$, instead of $\tilde{g}$, to populate $V$. Consequently, all elements corresponding to false positives (i.e., set $P - S$) receive the original, instead of zero values. In [45] we show that for a $\delta$-compressor $C_\delta$, there exists $\beta \in [0, 1)$, $\beta \geq \delta$ such that the compression error due to gradient $C_{P0,\delta}(g)$ resulted from policy P0, is $\mathbb{E}\|g - C_{P0,\delta}(g)\|^2 \leq (1-\beta)\|g\|^2$. In practical terms, we show in our experiments that the quality of the trained model when we use a Bloom filter with policy P0, is as good or better than that of the initial sparsifier.

**Random approach: Policy P1.** To address the increased data volume issue of policy P0, this policy defines a new set of indices $\tilde{S} \subseteq P$, where $|\tilde{S}| = r$. $\tilde{S}$ is generated by randomly selecting $r$ elements from $P$. Consequently, array $V$ is constructed as $\forall i \in [1, r] : V(i) = g[\tilde{S}[i]]$ (recall that GRACE allows access to the original gradient $g$). Since $\tilde{S} \neq S$ in general, we expected the error to be affected. Let $k_1 = |\tilde{S} \cap S|$ and assume that the input gradient is inherently sparse, i.e., sparsifier $C_\delta$ is the identify function $I_d$. For the combined compressor $C_{P1,I_d}$ with policy P1, we show [45] that $\mathbb{E}\|g - C_{P1,I_d}(g)\|^2 = (1 - \frac{k_1}{r})\|g\|^2$; that is, policy P1 creates a lossy compressor with compression factor as good as Random-$k_1$. In practice, DeepReduce allows $C_\delta$ to be any sparsifier (e.g., Top-$r$ [5, 8]). In this case, policy P1 is essentially equivalent to a combined sparsifier, similar to Elibol et al. [25] and Barnes et al. [9]. For a detailed analysis and proofs of error bounds, refer to our technical report [45].

**Conflict sets: Policy P2.** The previous policies represent two extremes: P0 eliminates errors but increases the amount of transfer data, whereas P1 sends fewer data but may introduce errors. Here, we propose policy P2, which transmits a modest amount of additional data compared to P1, and is close to P0 in terms of error. Similar to P1, this policy generates a set $\tilde{S} \subseteq P$, but makes better probabilistic choices. Intuitively, false positives are due to collisions in the Bloom filter, resulting in conflicts. Policy P2 groups all items of $P$ into conflict sets. Two elements $x$ and $y$ belong to a conflict set $C_j$ if $x, y \in P$ and $h_i(x) = h_{i'}(y) = j$ for $i, i' \in [k]$, where $j$ is the $j^{th}$ bit of $\mathcal{B}$ and $\bigcup_j C_j = P$. Figure 4 shows an example, with 4 items and 3 hash functions; item $i_7$ is a false positive and there exist 8 conflict sets. If a set $C_j$ contains only one item, it is guaranteed to be a true positive, so it is added to $\tilde{S}$. Else, if $C_j$ contains many items, a random subset is inserted into $\tilde{S}$. Therefore, it is possible that some false positives are included in $\tilde{S}$, but the probability is smaller, compared to policy P1.

Algorithm 1 presents the details. Lines 1-2 construct set $P$, i.e., the union of the true and false positive responses of $\mathcal{B}$. Lines 3-4 re-hash the items of $P$ into $\mathcal{B}$ to construct conflict

**Algorithm 1:** Construct bloom filter for policy BF-P2

**Input:** Bloom filter $\mathcal{B}$ of size $m$, $k$-hash functions $h_i$, gradient dimensionality $d$, empty set $P$, empty conflict sets $C_j$, empty set $\tilde{S}$, target number of decompressed indices $r$
**Output:** A set of decompressed indices $\tilde{S}$

1 **for** $i = 1$ *to* $d$ **do** /* all $d$ elements of gradient $\tilde{g} \in \mathbb{R}^d$ */
2      **if** $i \in \mathcal{B}$ **then** insert $i$ in $P$
3 **for** *each* $x \in P$ **do**
4      **for** $i = 1$ *to* $k$ **do** insert $x$ in $C_{h_i(x)}$
5 Sort conflict-sets in $C$ by their sizes in ascending order
6 **while** $size(\tilde{S}) < r$ **do**
7      **for** *each* $C_j \in C$ **do**
8          **if** $|C_j| = 1$ **then** Insert $C_j$ in $\tilde{S}$; Remove $C_j$ from $C$
9          **else**
10              Remove from $C_j$ items that exist in $\tilde{S}$
11              Insert into $\tilde{S}$ a random item from $C_j$
12 **return** $\tilde{S}$

sets $C_{1...j}$, where $j$ is equal to the number of "1"s in $\mathcal{B}$. For lack of better information, we assume that the true positives are uniformly distributed across the conflict sets; therefore, the probability of drawing a true positive out of a smaller conflict set is higher. To prioritize such sets, Line 5 sorts $C_{1...j}$ in ascending size order. Then, lines 6-11 repeatedly draw items out of the conflict sets until the size of $\tilde{S}$ reaches our target size $r$. If a set $C_j$ is initially a singleton, its item is a true positive; thus it is added to $\tilde{S}$. Else, we remove from $C_j$ any items that already exist in $\tilde{S}$ (observe there may be duplicates among conflict sets), and add randomly a remaining item to $\tilde{S}$.

**Implementation on GPUs and CPUs.** We provide an efficient GPU implementation of Bloom filters on PyTorch. During construction, many items can be inserted in parallel without locking, since collisions do not cause inconsistency. Since the domain $[d]$ of the hash functions is finite, we precompute a 2D lookup table $\mathbb{H}^{d,k}$, for each possible input of all hash functions. We store $\mathbb{H}$ in the GPU memory, allowing us to insert items in the Bloom filter using only lookup operations. $\mathbb{H}$ occupies around 1.5MB for ResNet-20 and 1GB for NCF; note that this optimization may not be feasible for very large models. Querying is also implemented in the GPU. If an item $i$ belongs to the Bloom filter, then $\mathcal{B}[h_1(i)] + \mathcal{B}[h_2(i)] + \mathcal{B}[h_3(i)] + \cdots + \mathcal{B}[h_k(1)] == k$. The summation can be executed in parallel with each hash function reduced to a lookup in $\mathbb{H}$. Moreover, many such queries can run concurrently. Although the basic Bloom filter is implemented on GPUs, complex policies, like P2, require programming flexibility. For this reason, we provide CPU implementations on PyTorch, using library pybloomfilter [3]; and on TensorFlow using the C++ extension to create custom operators.

## 5 Curve Fitting for values

We propose a novel curve fitting-based approach for the independent compression of gradient values. Figure 5 shows, in light blue, the gradient values for one layer of ResNet20 on CIFAR-10. By sorting those values, we obtain a smooth curve that can be approximated by a set of functions. This results to high compression ratios, even after considering the overhead of the mapping from the original to the sorted order.

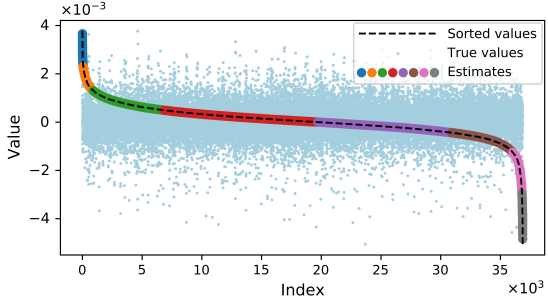

Figure 5: Piece-wise (8 pieces) value fitting on a convolution layer gradient of ResNet20 (on CIFAR-10).

Formally, let $g$ be the original gradient and $C_\delta(g)$ be the sparse gradient resulting from a $\delta$-sparsifier. We sort the values of the nonzero components of $C_\delta(g)$ in descending order and denote as $C_S(g)$. Let $g_t$ be the gradient estimate at iteration $t$ of a DNN training, and let $C_S(g_t) \in \mathbb{R}^d$ follow the hierarchical model: $C_S(g_t) = \nabla f_t + \sigma \xi_t$, where $\xi_t \sim N(0, I_d)$, $\nabla f_t \in \mathbb{R}^d$ is the sorted oracle gradient and $\sigma \in \mathbb{R}^+$; the model is commonly used in signal processing for approximation problems [4, 40, 66]. With these assumptions, we show how to fit regression models on $D := \{i, C_S(g_t)[i]\}_{i=1}^r$ and calculate the fitting error.

**Piece-wise approximation.** For piece-wise approximation, we split the sorted curve into segments and fit each segment individually. We explain how this is done for the positive sorted values; similar idea applies to the negative sorted values. Let the whole gradient be sorted in descending manner; set $[l]$ corresponds to indices of the sorted positive values and $[d] \setminus [l]$ to the sorted negative values. Let $y = mx + c$ be the line joining $(1, C_S(g)[1])$ and $(l, C_S(g)[l])$. Calculate $d_i := (y_i - C_S(g)[i])^2$, where $y_i = mi + c$. Choose the sorted gradient component that corresponds to $\max_{i \in [l]} d_i$ as the segmentation point. The process continues until the desired number of segments is reached. We stop segmentation if the number of points in a segment is less than $(n' + 1)$, where $n'$ is the degree of the polynomial used to fit on each segment.

**Polynomial regression.** Over each segment, we apply polynomial regression. As it is well-known (cf. [17]), piece-wise polynomial regression (least squares approximation) with free knots is a difficult theoretical problem and is of current research interest. Although there exist algorithms to numerically find the optimal solution, most are computational intensive which makes them not suitable for our use at each iteration of DNN training. We thus propose a simplified way of segmentation (as described before and then follow it by a regression). Due to limited space, we focus on the piece-wise linear case. For result concerning piece-wise constant approximation, see our technical report [45]. For piece-wise linear fit, we provide the following result with explicit constants.

**Lemma 1.** *(Error of piece-wise linear fit) For $C_S(g) \in C^1([1, d])$ with $\mathrm{Var}_{[1,d]}(C_S'(g)) \leq M$, we have $\|s - C_S(g)\|_\infty \leq \frac{2M}{p^2}$, for some $s \in S_p^1$—the set of piece-wise linear splines with $p$ knots.*

Since the least squares error will be less than $\sqrt{d}\|s - C_S(g)\|_\infty$, we see that the error can be controlled by taking large $p$. We give a heuristic to calculate $p$ from Lemma 1. First, calculate $M = |(C_S(g)[1] - C_S(g)[2]) - (C_S(g)[d-1] - C_S(g)[d])|$. By considering the error bound $\frac{2M}{p^2}$ as a function of $p$, we find closed-form solution for $p$ as $p = \lceil 2\sqrt{M} \rceil$.

**Nonlinear regression.** It is possible to use nonlinear regression for value fitting, such as double exponential model, $y = ae^{bx} + ce^{dx}$, where $(a, b, c, d) \in \mathbb{R}^4$ are the parameters. We empirically show this in Section 6.

**Implementation.** The piece-wise polynomial regression can be solved as a linear problem, once the segments are determined. Our GPU implementation uses Least-Square fitting, which can be trivially expressed with tensor operations. We also provide a CPU implementation using polyfit from the NumPy [54] library. We implement the nonlinear double exponential regression on TensorFlow, using tensor operations.

## 5.1 Combined index and value compression

To combine Bloom filter-based with curve fitting-based compression, first observe that neither method is order preserving. Therefore, we need a mapping from the original to the final position of each value. This corresponds to a 1D vector with $1 \sim d$, where $d$ is the size of sparse gradient. Since now the maximum element in this mapping vector is $d$, we encode their each element using $\lceil \log_2(d) \rceil$ bits. For our experiments, this corresponds to 16 bits for ResNet50 and 19 bits for NCF, which is a significant gain compared to the usual int32 format.

## 6 Experimental Evaluation

**Implementation.** DeepReduce supports TensorFlow and Pytorch. We provide various versions of our index and value compressors, on CPUs as well as GPUs. We also instantiate our framework with combinations of existing methods, namely Huffman and RLE for index compression, as well as Deflate and QSGD [7] for value compression; see our technical report [45] for a list. We denote implementations that use our framework by $\mathrm{DR}_{\mathrm{idx}}^{\mathrm{val}}$, where *idx* and *val* is the index and value compression method, respectively. We compare against state of the art gradient compression approaches, namely: 3LC [47], SketchML [38] and SKCompress [39].

**Environment.** We use 8 dedicated machines with Ubuntu 18.04.2 LTS and Linux v.4.15.0-74, 16-Core Intel Xeon Silver 4112 @ 2.6GHz, 512 GB RAM, one NVIDIA Tesla V100 GPU card with 16 GB on-board memory and 100Gbps network. For time-insensitive metrics (e.g., accuracy, data volume), we also use a shared cluster with NVIDIA Tesla V100 GPUs with 32GB on-board memory. We deploy CUDA 10.1, TensorFlow 1.14, PyTorch 1.4.0, Horovod 0.18.2, OpenMPI 4.0 and NCCL 2.4.8.

**Benchmarks.** We use industry-standard benchmarks from TensorFlow [50, 67] and NVIDIA [53] (see Table 1). The benchmarks span two common deep learning tasks: image classification and recommendation; the training quality is measured by top-1 accuracy and best hit rate, respectively. The number of communicated gradient vectors ranges from 10 to 161.

Table 1: Summary of the benchmarks and datasets; last collumn shows the best quality achieved by the no-compression baseline.

| Type | Model | Task | Dataset | Parameters | Optimizer | Platform | Metric | Baseline |
|---|---|---|---|---|---|---|---|---|
| CNN | ResNet-20 [42] | Image classif. | CIFAR-10 [46] | 269,722 | SGD-M | TFlow | Top-1 Acc. | 90.94% |
| | DenseNet40-K12 [35] | Image classif. | CIFAR-10 [46] | 357,491 | SGD-M | TFlow | Top-1 Acc. | 91.76% |
| | ResNet-50 [42] | Image classif. | ImageNet [18] | 25,557,032 | SGD-M | TFlow | Top-1 Acc. | 73.78% |
| MLP | NCF [32] | Recommendation | Movielens-20M [52] | 31,832,577 | Adam | PyTorch | Best Hit Rate | 94.97% |

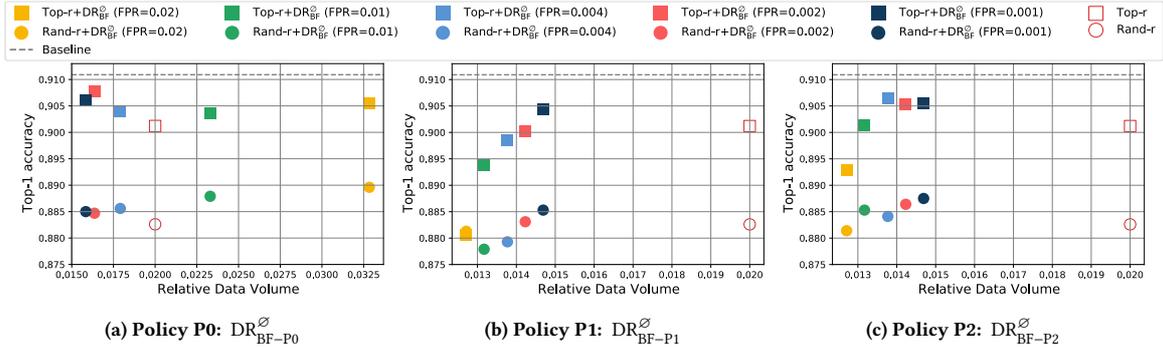

(a) Policy P0: $\mathrm{DR}^{\varnothing}_{\mathrm{BF-P0}}$

(b) Policy P1: $\mathrm{DR}^{\varnothing}_{\mathrm{BF-P1}}$

(c) Policy P2: $\mathrm{DR}^{\varnothing}_{\mathrm{BF-P2}}$

Figure 6: Effect of FPR on top-1 accuracy for the three Bloom filter policies, for ResNet-20 on CIFAR-10. The sparse input gradients were generated by Top-$r$ and Random-$r$ sparsification methods. Data volume is relative to the no-compression baseline.

## 6.1 Bloom filter-based index compression

**Effect of false positive rate (FPR).** We train ResNet-20 on CIFAR-10 on 8 nodes for 328 epochs and measure the top-1 accuracy and the amount of transferred data. Our baseline transmits the original uncompressed gradients. To generate sparse gradients, we employ the Top-$r$ [6] and Rand-$r$ [63] sparsifiers; each achieves different accuracy [75]. We vary FPR and measure its effect; smaller FPR corresponds to larger bloom filter. The results for our three index compression policies are shown in Figure 6. Recall (Section 4) that policy BF-P0 transmits extra data for each false positive index. The advantage, as shown in Figure 6a is that accuracy is only marginally affected by FPR, irrespectively of the gradient sparsifier (i.e., Top-$r$ or Rand-$r$). The disadvantage is that the amount of transferred data increases with higher FPR; if it is high enough (e.g., more than 0.004 in our figure), then BF-P0 transfers more data than the sparse input gradient. Policy BF-P1, on the other hand, resolves bloom filter conflicts randomly; as expected, Figure 6b confirms that the amount of transferred data decreases when FPR increases. The trade-off is that accuracy also decreases because more erroneous gradient elements are received. Fortunately, our next policy, BF-P2, improves this issue, as shown in Figure 6c; by resolving conflicts in an informed way. In a sense, BF-P2 combines the benefits of BF-P0 and BF-P1, while avoiding their pitfalls.

**Convergence rate of each policy.** Figure 7 depicts the convergence timeline for our three index compression polices for ResNet-20 on CIFAR-10; FPR is set to 0.001. We compare against the no-compression baseline, as well as the plain Top-$r$ sparsifier. All our policies converge to the same top-1 accuracy as the no-compression baseline, but BF-P2 converges in fewer training epochs. It is worth noting that BF-P2 converges faster than the plain Top-$r$ sparsifier, despite transmitting 33% fewer data (refer to Figure 6c). We also compare against BF-naïve (Section 4), achieves much lower accuracy, justifying the need for our proposed policies. Similar results were observed for DenseNet40-K12; refer to our technical report [45].

## 6.2 Curve fitting-based value compression

Figure 8 shows the converge timeline for our two curve fitting-based value compression methods, Fit-Poly and Fit-DExp, on ResNet-20 with CIFAR-10. The sparse input gradients were generated by Top-$r$. Fit-Poly uses polynomials with degree 5 (i.e., 6 coefficients).Fit-DExp requires 4 coefficients without segmentation. Both methods converge to the same accuracy as the no-compression baseline; also, both converge in fewer steps than plain Top-$r$. Fit-DExp has a slight advantage over Fit-Poly, in terms of convergence. Moreover Fit-DExp sends fewer data: it compresses the output of Top-$r$ by roughly 50%, whereas Fit-Poly compresses it slightly less, by around 40%. The trade-off is the compression and decompression overhead: Fit-DExp is roughly 3.5× slower than Fit-Poly. Details about the data volume and runtime are discussed in Figure 10 below.

## 6.3 DeepReduce vs stand-alone methods

DeepReduce is designed to work either in conjunction with a sparsifier (e.g., Top-$r$), or with inherently sparse gradients. To contrast with the common practice, we compare DeepReduce against state-of-the-art *stand-alone* gradient compressors [75] that are applied directly on the original gradient.

**DeepReduce on top of the Top-$r$ sparsifier.** We consider two instantiations of DeepReduce: (*i*) $\mathrm{DR}^{\varnothing}_{\mathrm{BF-P2}}$ that uses Bloom filter index compression with policy P2 and $FPR = 0.001$; and (*ii*) $\mathrm{DR}^{\mathrm{Fit-Poly}}_{\varnothing}$ that uses value compression with polynomial fit. Both operate on the sparse tensor generated by Top-$r$, with $r = 1\%$. We compare against two stand-alone gradient compressors: 3LC [48] with spasification multiplier set to 1, and SketchML [37]; we use $2^6$ quantile buckets, since we opt for best accuracy. Memory compensation is enabled for all methods. For this experiment, we employ a much larger benchmark: ResNet-50 on ImageNet. The results are shown in Figure 9, where data volume is relative to the no-compression baseline. Both DeepReduce instantiations provide a good balance between data volume and accuracy, whereas each of the stand-alone methods is biased towards one of the two metrics.

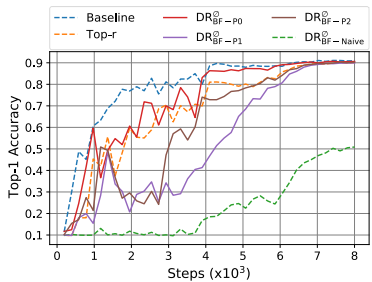 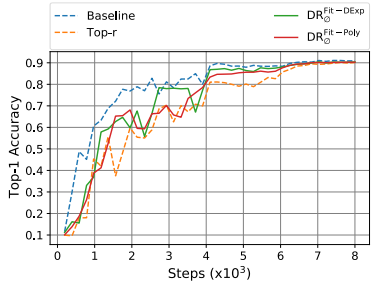 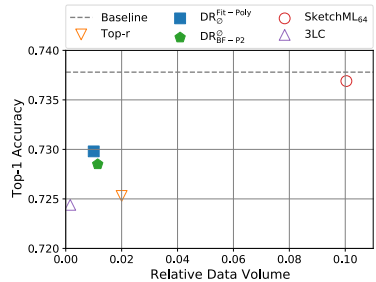

Figure 7: Convergence timeline for our bloom filter policies for ResNet-20 on CIFAR-10; FPR = 0.001. $\mathrm{DR}_{\mathrm{BF-P2}}^{\varnothing}$ is the fastest. Accuracy suffers for $\mathrm{DR}_{\mathrm{BF-Naive}}^{\varnothing}$.

Figure 8: Convergence timeline of value compressors, for ResNet-20 on CIFAR-10. $\mathrm{DR}_{\varnothing}^{\mathrm{Fit-DExp}}$ is slightly better than $\mathrm{DR}_{\varnothing}^{\mathrm{Fit-Poly}}$.

Figure 9: DeepReduce with Top-$r$ against stand-alone compressors, for ResNet-50 on Imagenet. Data volume is relative to the no-compression baseline.

**DeepReduce for an inherently sparse model.** Language and recommendation models are often sparse; for example, the gradients of NCF consist of roughly 40% zeros. For such cases, DeepReduce can be applied directly without any sparsifier. In this experiment, we train NCF on ML-20m, with $10^6$ local batch size. We test two instantiations of DeepReduce: (*i*) $\mathrm{DR}_{\mathrm{BF-P2}}^{\mathrm{Fit-Poly}}$, which combines BF-P2 (FPR=0.01) for indices with Fit-Poly for values; and (*ii*) $\mathrm{DR}_{\mathrm{BF-P0}}^{\mathrm{QSGD}}$, which uses BF-P0 (FPR=0.6) for indices, but combines it with QSGD [7], an existing method for value compression. This demonstrates that DeepReduce is compatible with various existing compressors. We compare against SKCompress [39], an improved version of SketchML (see above), optimized for sparse tensors. We configure QSGD and SKCompress for 7-bits quantization and set the QSGD bucket size to 512. For SKCompress, we omit the grouped MinMaxSketch and separation of positive/negative gradients, as they have only minor effects. Table 2 shows the results. All methods achieve virtually the same best hit rate (i.e., the quality metric for NCF), but $\mathrm{DR}_{\mathrm{BF-P0}}^{\mathrm{QSGD}}$ is better than $\mathrm{DR}_{\mathrm{BF-P2}}^{\mathrm{Fit-Poly}}$ in terms of data volume, because the latter transmits additional mapping information for the re-arranged elements (see Section 5). $\mathrm{DR}_{\mathrm{BF-P0}}^{\mathrm{QSGD}}$ is also better than SKCompress. Although the improvement appears small, $\mathrm{DR}_{\mathrm{BF-P0}}^{\mathrm{QSGD}}$ is easily implemented on GPUs; therefore, in our experiments (see Figure 10b) it is 380× faster in terms of compression and decompression time.

### 6.4 Practical applicability of DeepReduce

Any operation on the gradient imposes computational overheads that may exceed the benefits of the reduced data volume and affect the practical applicability, as discussed below.

**Data volume and computational overhead.** In Figure 10a, we show the data volume (relative to the no-compression baseline) separately for values and indices, for various instantiations of DeepReduce. We test Resnet-20 on CIFAR-10 and generate sparse tensors by Top-$r$. We compare against SKCompress, which also operates on the sparse tensor. For fairness, parameters are selected such that all methods achieve similar accuracy. Although the ratio of index over value data volume differs for each combination of DeepReduce, all versions transmit fewer data than plain Top-$r$. Interestingly, SKCompress performs the best; however, this depends on the particular model, as can be verified by the NCF results in Table 2.

Table 2: Inherently sparse model: NCF on ML-20m. Data volume is relative to baseline. $\mathrm{DR}_{\mathrm{BF-P0}}^{\mathrm{QSGD}}$ performs the best.

| Method | Relative Data Volume | Best Hit Rate |
| --- | --- | --- |
| Baseline | 1 | 0.9497 |
| $\mathrm{DR}_{\mathrm{BF-P2}}^{\mathrm{Fit-Poly}}$ | 0.5879 | 0.9519 |
| SKCompress | 0.2175 | 0.9513 |
| $\mathrm{DR}_{\mathrm{BF-P0}}^{\mathrm{QSGD}}$ | **0.2063** | 0.9496 |

Figure 10b shows, in logarithmic scale, the computational overhead of compression and decompression measured by the wall clock runtime. We implement all methods by utilizing the best available libraries either on CPUs or GPUs; we acknowledge there remains margin for improvement. Nonetheless, this experiment demonstrates the significant variation in terms of overhead among methods; for instance, SKCompress is 3 orders of magnitude slower, compared to $\mathrm{DR}_{\varnothing}^{\mathrm{QSGD}}$.

**Total training runtime.** We demonstrate a realistic deployment by training NCF on ML-20m and measuring the total wall clock time. We employ gradient accumulation with 10 accumulations per iteration and $10^6$ local batch size. We use NCCL Allreduce for baseline communication, and NCCL Allgather for Top-$r$ ($r = 10\%$) and DeepReduce. We vary the network bandwidth from 100Mbps up to 10Gbps, and consider 32-bit and 16-bit training precision. Figure 11 shows the wall time, in three components: forward and back-propagation, encoding / decoding, and communication. Gradient compression is beneficial only when the ratio of communication over computation cost is high; this is consistent with the findings in [51, 55, 75].

**Discussion.** The previous experiment demonstrates that the practical benefit of any compressor, including DeepReduce, depends on multiple factors that affect the communication over computation ratio. Those include the communication library (e.g., NCCL, Gloo), the implementation details of each compressor, the computing hardware (e.g., faster GPUs), possible accelerators (e.g., NICs with FPGAs), and others. Some applications are more suitable for gradient compression, such as Geo-distributed training [34] and Federated Learning [43], where network bandwidth is typically limited. The advantage of DeepReduce lies in its versatility to intermix various index and value compressors, e.g., it can combine a GPU implementation of QSGD (value quantization) with a streaming FPGA implementation of Bloom filter-based index compression.

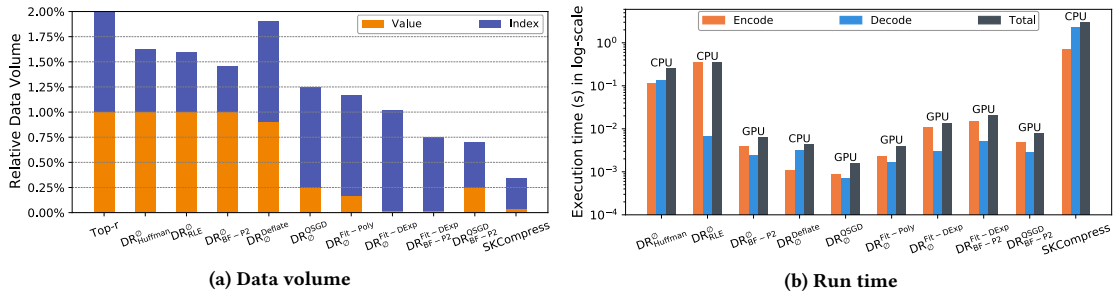

(a) Data volume  (b) Run time

Figure 10: Comparing various compression methods on the Top-$r$(1%) values of a convolution gradient in ResNet-20 (gradient size: 36864). (a) Data volume (b) Encoding and decoding runtime

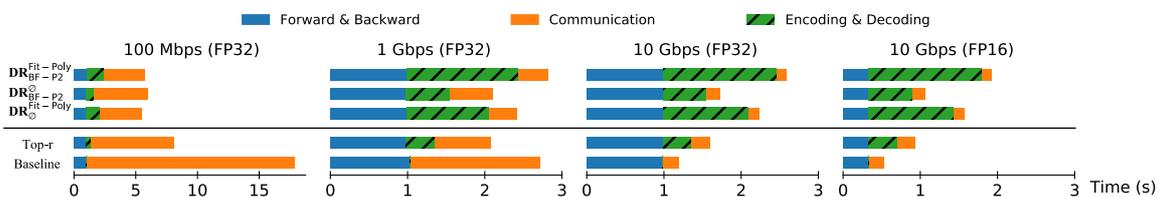

Figure 11: Time breakdown in one iteration training of NCF on ml-20m. We show the speedup of training by DR on 4 nodes with different network bandwidth: 100Mbps vs. 1Gbps vs. 10Gbps, and also with FP16 mixed precision training.

## 7 Related Work

**SketchML and SKCompress.** In SketchML [37], the nonzero gradient elements are quantized into buckets using a non-uniform quantile sketch. The number of buckets is further reduced via hash tables that resolve collisions by a Min-Max strategy. The indices are compressed by a delta encoder. SKCompress [39] improves SketchML by additional Huffman coding on the bucket indices as well as the prefix of delta keys. Both of these methods can be viewed as special cases of DeepReduce.

**Hybrid compressors.** Qsparse local SGD [10] combines quantization with Top-$r$ or Random-$r$ sparsifiers. Strom et al. [64] and Dryden et al. [23] use a fixed and adaptive threshold, respectively, to sparsify. Elibol et al. [25] combine Top-$r$ with randomized unbiased coordinate descent. Barnes et al. in [9], perform a Top-$m$ selection of each local gradient and communicate $r < m$ randomly chosen components. Double quantization [76] is an asynchronous approach that integrates gradient sparsification with model parameter and gradient quantization. The output sparse gradient of hybrid methods can be the input to our framework; therefore, our work is orthogonal.

**Quantization and encoding.** Quantized gradients require efficient encoding. QSGD [7] and TernGrad [73] use Elias encoding, while Natural Compression [33] uses fixed length 8-bit codes. Gajjala et al. [28] use Huffman encoding, whereas Faghri et al. [26] use adaptive quantization with 3 bits.

**Sparse tensor communication.** Communication libraries typically transmit sparse tensors via Allgather [1], because the more efficient Allreduce collective only supports dense tensors. In contrast, ScaleCom [14] tailors Allreduce to sparse data. OmniReduce [27] also implements sparse Allreduce that sends the non-zero blocks to the workers in an all-to-all manner. SparCML [57] adaptively switches between Allreduce and Allgather based on global gradient sparsity among the workers. SwitchML [60] is a hardware approach that aggregates the model updates in programmable network switches.

## 8 Conclusions

Sparse tensors are ubiquitous in distributed DNN training. DeepReduce integrates seamlessly with popular machine learning frameworks and provides an easy-to-use API for the effortless implantation of a wide variety of sparse tensor communication methods. We demonstrate the practical applicability of DeepReduce both by instantiating it with existing index and value compressors, as well as by implementing two novel methods: a Bloom filter-based index compressor and a curve fitting-based value compressor. DeepReduce targets large scale models, and can assist practitioners to lower the communication overhead for their specific distributed training pipeline.

## Acknowledgments

Kelly Kostopoulou was supported by the KAUST Visiting Student Research Program. The computing infrastructure was provided by the KAUST Super-computing Lab (KSL).

# 9 Appendix
## 9.1 Background

Figure 12 shows an example of distributed training for DNNs with compressed communication.

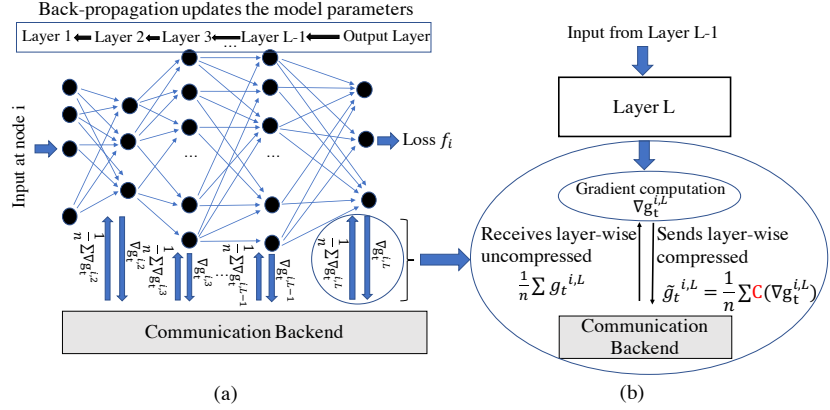

Figure 12: Distributed training from the perspective of $i^{\text{th}}$ computing node.

## 9.2 Bloom filter-based index compression

In this scope we explain Naive compression by a simple illustration.

**Naive compression.** The process we described is illustrated in Figure 13. We have a dense gradient represented as a sparse tensor in a key-value format. Notice that, the values in the sparse representation are sorted by their indices in an increasing order. The set, $S$ of keys is represented as a bloom filter. To communicate the sparse tensor we send both the values and the bloom filter. During the phase of decompression, we try to reconstruct $S$ by following the process we described. However, in this case, we manage to retrieve only 4 out of the 5 elements of $S$. Index 4 does not belong to $S$ and corresponds to a FP response. Mapping $\mathcal{M}$ scans the communicated values in the order they arrive and assigns each one of them to the next larger index from the set of decoded indices. Notice how the selection of one wrong index affects the decompression by causing re-arrangements or shifts of the reconstructed gradient components with respect to their true positions.

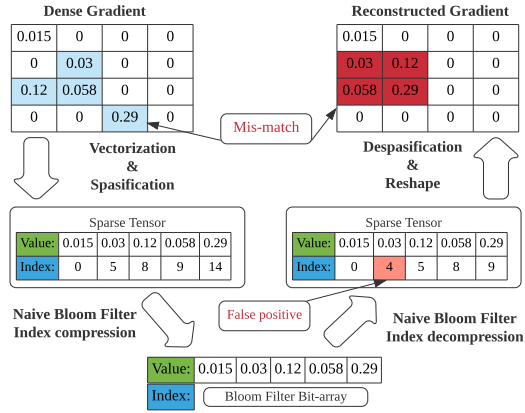

Figure 13: Bloom Filter—Naive Bloom Error

## 10 Theoretical Insights

### 10.1 Inequalities used

(1) If $a, b \in \mathbb{R}^d$ then the Peter-Paul inequality is: There exists a $\xi > 0$ such that

$$\|a + b\|^2 \leq (1 + \xi)\|a\|^2 + (1 + \frac{1}{\xi})\|b\|^2. \tag{1}$$

We generally use a relaxed version of the above inequality as follows:

$$\|a + b\|^2 \leq 2\|a\|^2 + 2\|b\|^2. \tag{2}$$

(2) If $a, b \in \mathbb{R}^d$ then there exists a $\xi > 0$ such that

$$2\langle a, b \rangle \leq \xi \|a\|^2 + \frac{1}{\xi}\|b\|^2. \tag{3}$$

(3) For $x_i \in \mathbb{R}^d$ we have:

$$\|\sum_{i=1}^n x_i\|^2 \leq n \sum_{i=1}^n \|x_i\|^2. \tag{4}$$

(4) If the operator $C : \mathbb{R}^d \to \mathbb{R}^d$ is a *compressor* then there exists $\Omega > 0$ such that

$$\mathbb{E}\|g - C(g)\|^2 \leq \Omega\|g\|^2. \tag{5}$$

### 10.2 Preliminaries

**Random-$r$, Top-$r$ sparsifiers, and $\delta$-compressors.** Based on the selection criteria of the elements in $S$, some of the most commonly used sparsifiers such as, Random-$r$ [63] and Top-$r$ [5, 8] are defined. For Random-$r$, the elements of $S$ are randomly selected out of $[d]$, whereas, for Top-$r$, the elements of $S$ correspond to the indices of the $r$ highest magnitude elements in $g$. Sparsifiers that follow (5) with $\Omega = 1 - \delta$, and $\delta \in (0, 1]$, are known as $\delta$-*compressors* and denoted by $C_\delta$. That is,

$$\mathbb{E}\|g - C_\delta(g)\|^2 \leq (1 - \delta)\|g\|^2. \tag{6}$$

*Remark* 1. Both Top-$r$ and Random-$r$ are $\delta$-compressors with $\delta = \frac{r}{d}$, and $\mathbb{E}\|g - \text{Top}r(g)\|^2 \leq \mathbb{E}\|g - \text{Random}r(g)\|^2 = (1 - r/d)\|g\|^2$, for all $g \in \mathbb{R}^d$.

The next Lemma characterizes the probability of the false-positive rates in a Bloom filter.

**Lemma 2.** *[12] Let $k$ denote the number of independent hash functions, $m$ the dimension of the bit-string, and $r$ the cardinality of the index set, $S$. Then the probability of the false-positive rate (FPR) is $\epsilon \approx (1 - e^{-kr/m})^k$.*

*Remark* 2. Given $\epsilon$ and $r$, the optimal $m = -\frac{r \log \epsilon}{(\log 2)^2}$ and $k = -\frac{\log \epsilon}{\log 2}$. Given $m$ and $r$, the number of hash functions that minimizes the probability of false positives is $k = \frac{m}{r} \log 2$. This $k$ results in the probability of false positive, $\epsilon$ as $\log \epsilon = -\frac{m}{r}(\log 2)^2$. In practice, we need to calculate the bits in the filter by using the relation $m = \frac{-r \log \epsilon}{(\log 2)^2}$ and the number of hash functions by $k = \frac{-\log \epsilon}{\log 2}$.

The next two Lemmas are instrumental in proving other compression related results.

**Lemma 3.** *Let $x \in \mathbb{R}^d$ and $x_S$ be a vector that has the components of $x$ arranged in ascending/descending order of magnitude. If $0 \leq \theta < \pi/2$ be the angle between $x$ and $x_S$, then $\|x - x_S\| = 2(1 - \cos\theta)\|x\|^2$.*

Proof. We have

$$\|x - x_S\|^2 = \|x\|^2 + \|x_S\|^2 - 2\langle x, x_S \rangle \stackrel{\|x\|=\|x_S\|}{=} 2\|x\|^2 - 2\|x\|^2 \cos\theta = 2(1 - \cos\theta)\|x\|^2.$$

Hence the result. □

**Lemma 4.** *Let $C(\cdot) : \mathbb{R}^d \to \mathbb{R}^d$ be a $\delta$-compressor.*
*(i) If $C_\delta(g)$ is unbiased then $\mathbb{E}\|C_\delta(g)\|^2 \leq (2 - \delta)\|g\|^2$.*
*(ii) If $C_\delta(g)$ is biased then $\mathbb{E}\|C_\delta(g)\|^2 \leq 2(2 - \delta)\|g\|^2$.*

Proof. Recall from (6), for $\delta$-compressors, we have $\mathbb{E}\|g - C_\delta(g)\|^2 \leq (1 - \delta)\|g\|^2$. Therefore,

$$\mathbb{E}\|C_\delta(g)\|^2 = \mathbb{E}\|g - g + C_\delta(g)\|^2 \stackrel{\mathbb{E}(C_\delta(g))=g}{=} \mathbb{E}\|g - C_\delta(g)\|^2 + \|g\|^2 \stackrel{\text{By (6)}}{\leq} (1 - \delta)\|g\|^2 + \|g\|^2 = (2 - \delta)\|g\|^2.$$

On the other hand, for biased compressors, $\mathbb{E}(C_\delta(g)) \neq g$ and hence,

$$\mathbb{E}\|C_\delta(g)\|^2 = \mathbb{E}\|g - g + C_\delta(g)\|^2 \leq 2\mathbb{E}\|g - C_\delta(g)\|^2 + 2\|g\|^2 \stackrel{\text{By (6)}}{\leq} 2(1 - \delta)\|g\|^2 + 2\|g\|^2 = 2(2 - \delta)\|g\|^2.$$

□

## 10.3 No Policy Error

**Proof of Lemma 5**

**Lemma 5.** *The cardinality of the set $P$ is at most $\lceil r + (\frac{1}{2})^{-\frac{\log(\epsilon)}{\log(2)}}(d-r)\rceil$ and approaches to $r$ as $\epsilon \to 0$.*

Proof. For given $\epsilon, r, d$ the cardinality of the set of positives $P$ follows

$$r \leq |P| \leq \lceil r + \epsilon(d-r) \rceil \stackrel{\text{By Lemma 2}}{\approx} \lceil r + (1 - e^{-kr/m})^k(d-r) \rceil.$$

Given $\epsilon$ and $r$, the optimal $m = -\frac{r \log \epsilon}{(\log 2)^2}$ and $k = -\frac{\log \epsilon}{\log 2}$. Therefore, plugging them in the above expression we get

$$r \leq |P| \leq \lceil r + \left(\frac{1}{2}\right)^{-\frac{\log(\epsilon)}{\log(2)}}(d-r) \rceil,$$

which after taking limit $\epsilon \to 0$ gives the desired result. □

The next Lemma measures compression error due to compressed gradient $C_{\mathcal{P}_0, \delta}(g)$.

**Lemma 6.** *(i) For inherently sparse gradient, $g$, with $C_\delta = I_d$, we have $\beta = \delta = 1$.*
*(ii) For a general $\delta$-compressor, $C_\delta$, there exists a $\beta \in [0, 1)$, $\beta \geq \delta$ such that the compression error due to a compressed gradient $C_{\mathcal{P}_0, \delta}(g)$ resulted from $\mathcal{P}_0$ is $\mathbb{E}\|g - C_{\mathcal{P}_0, \delta}(g)\|^2 \leq (1-\beta)\|g\|^2$.*

Proof. (i) For inherently sparse gradient, $g$, with $C_\delta = I_d$, we have $\delta = 1$, and $\|C_{I_d}(g)\|_0 = r$. For policy $P_0$, we have $\|C_{P_0, I_d}(g)\|_0 = r$ resulting to $\beta = \delta = 1$.

(ii) Consider a $\delta$ sparsifier, $C_\delta$ such that $\|C_\delta(g)\|_0 = r$. If $P_0$ is used then, by Lemma 5, $|P| = r + (\frac{1}{2})^{-\frac{\log \epsilon}{\log 2}}(d-r) \geq r$ for $\epsilon \geq 0$, ($|P| = r$ for $\epsilon = 0$). Therefore, for $C_{P_0, \delta}$, we have $\beta = |P|/d > \delta$ resulting $\mathbb{E}\|g - C_{\mathcal{P}_0, \delta}(g)\|^2 \leq (1-\beta)\|g\|^2$. □

*Remark* 3. We consider two extreme cases of $\delta$ sparsifier. For Random-$r$, $\delta = r/d \leq 1$. If $P_0$ is used then $|P| = r + (\frac{1}{2})^{-\frac{\log \epsilon}{\log 2}}(d-r) > r$ for $\epsilon > 0$. In this case, for $C_{P_0, \delta}$ we have $\beta = |P|/d > \delta$.

In another extreme case, for $C_\delta$ to be Top-$r$, by Remark 1, $\delta \leq r/d \leq 1$. For $g \in \mathbb{R}^d$, similar argument as above gives us:

$$\mathbb{E}\|g - C_{P_0, \delta}\|^2 < \mathbb{E}\|g - \text{Top}r(g)\|^2 \leq \mathbb{E}\|g - \text{random}r(g)\|^2 \leq (1 - r/d)\|g\|^2.$$

Lemma 5 show that for small $\epsilon$, no policy sends negligible amount of extra data compared to the other policies. For inherently sparse tensors, Lemma 6 (*i*) shows that no policy is *lossless* and is the best choice. In contrast, Lemma 6 (*ii*) shows that for sparsified vectors, no policy achieves a better compression factor than the original sparsifier, $C_\delta$.

## 10.4 Deterministic policy

This policy deterministically selects a subset of $r$ elements from $P$ and is denoted by $\mathcal{P}_D$. One can select the first $r$, the middle $r$, or the last $r$ elements from $P$, and based on this denote them as, leftmost-$r$, middle-$r$, and rightmost-$r$ policy, respectively. For implementation, the set $\tilde{S}$ can be created while iterating and posing queries on the universe $U$ and once it has $r$ elements, the querying is stopped. Let $C_\delta$ be a general $\delta$-compressor that selects $r$ gradient components. However, with a policy $\mathcal{P}_D$, not all the $r$ selected indices are due to $C_\delta$. Let $I_1$ denote the set of indices that are selected via policy $\mathcal{P}_D$ originally resulted from $C_\delta$ sparsifier and let $I_2$ denote the set of the rest of the $(r - |I_1|)$ indices. Therefore, $\tilde{S} = I_1 \bigcup I_2$ and let $C_{\mathcal{P}_D, \delta}(g)$ be the compressed gradient whose indices are drawn via policy $\mathcal{P}_D$ and has support $\tilde{S}$. The following lemma quantifies the compression error.

**Deterministic policy error.** Lemma 7 gives the compression error bound for Deterministic policies.

**Lemma 7.** *(i) For all deterministic policies, $\mathcal{P}_D$ and for an inherently sparse gradient $g$, the compression error due to a compressed gradient, $C_{\mathcal{P}_D, I_d}(g)$ is given by $\mathbb{E}_{\tilde{S}}\|g - C_{\mathcal{P}_D, I_d}(g)\|^2 = (1 - \frac{|I_1|}{r})\|g\|^2$.*
*(ii) For all deterministic policies, $\mathcal{P}_D$ and a general $C_\delta$, the compression error due to a compressed gradient, $C_{\mathcal{P}_D, \delta}(g)$ is given by $\mathbb{E}_{\tilde{S}}\|g - C_{\mathcal{P}_D, \delta}(g)\|^2 = (1 - \frac{r}{d})\|g\|^2$.*

The proof of Lemma 7 follows from standard procedure of taking expectation with respect to all possible $|I_1|$ cardinality subsets formed from the set $P$ by policy $P_D$ and follows the same structure as the proof of Lemma 8 (*i*) and (*ii*). We omit the proof.

Lemma 7 (*i*) shows that by adopting a deterministic policy, $\mathcal{P}_D$ on the support of a general $\delta$-compressor, the compression error in expectation is as good as using a Random-$r$ sparsifier on the original gradient. Moreover, Lemma 7 (*ii*) shows that by adopting a deterministic policy, $\mathcal{P}_D$ on the support of a inherently sparse vector, the compression error in expectation is as good as using a Random-$|I_1|$ sparsifier on the original gradient. Therefore, it creates a *lossy* compression whose compression factor is as good as a Random-$|I_1|$ compressor on $r$ elements.

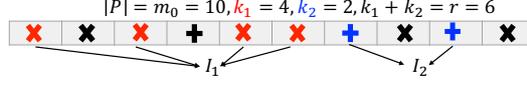

Figure 14: Random Policy example with ✖ as TP and ➕ as FP. A random policy selects set $I_1$ with ✖ and $I_2$ with ➕.

## 10.5 Random Policy

**Random approach: Policy P1.** A deterministic policy, $\mathcal{P}_D$ may be prone to bias, based on how the gradient components are distributed and the sparsifier used. E.g. If the support set of the sparsifier $C_\delta$ is concentrated at the beginning of $P$, then except leftmost-$r$, other deterministic policies such as, middle-$r$ and rightmost-$r$ incur more bias as they would select more elements in $I_2$ than $I_1$. Similarly, if the support set of the sparsifier $C_\delta$, concentrated at the center, then only middle-$r$ policy is expected to incur the least bias compare to the others as it selects more elements in $I_1$ than $I_2$. Without knowing the distribution of the gradient components, it is hard to invoke a deterministic policy. A random policy, $\mathcal{P}_R$ forms $\tilde{S}$ by picking $r$ indices randomly from $P$. Without loss of generality, consider the only source of randomness here is due to the random selection of $r$ indices and is unaffected by other source of randomness—randomness in the i.i.d. data, independence of the hash functions, etc.

Let $|P| = m_0$. Let policy $\mathcal{P}_R$ chooses a set $I_1$ of $k_1$ elements from the support set of a $\delta$-compressor $C_\delta$ without replacement. Let the rest $k_2$ elements belong to the set $I_2$ such that, $k_1 + k_2 = r$. We illustrate this in Figure 14. By this, we incur two types of errors with respect to the vector $g_P$. The first error, $E_1$, is due to the compressed gradient $g_{I_1}$. The second error, $E_2$, is due to the compressed gradient $g_{I_2}$. We have, $C_{\mathcal{P}_R,\delta}(g) := g_{I_1} \bigoplus g_{I_2}$. In the following Lemma, we measure the compression error due to compressed gradient $C_{\mathcal{P}_R,\delta}(g)$ with respect to $g_P$.

**Lemma 8.** *With the notations mentioned above, we have the following measures of the compression error:*
*(i) $E_1 = \mathbb{E}_{\mathcal{P}_R} \|g_P - g_{I_1}\|^2 = (1 - \frac{k_1}{r})\|g_P\|^2$.*
*(ii) $E_2 = \mathbb{E}_{\mathcal{P}_R} \|g_P - g_{I_2}\|^2 = (1 - \frac{k_2}{m_0-r})\|g_P\|^2$.*
*(iii) Denote $E := \mathbb{E}_{\mathcal{P}_R} \|g_P - C_{\mathcal{P}_R,\delta}(g)\|^2$ be the total compression error due to the compressed gradient $C_{\mathcal{P}_R,\delta}(g)$ with respect to $g_P$. Then, $E \leq E_1 + E_2$.*

PROOF. Let $\Omega_{k_1}$ and $\Omega_{k_2}$ denote the set of all $k_1$ and $k_2$ elements subsets of the sets having cardinality $r$ and $m_0 - r$, respectively. The first error, $E_1$, is due to the compressed gradient $g_{I_1}$ whose support belongs to $\Omega_{k_1}$. The second error, $E_2$, is due to the compressed gradient $g_{I_2}$ whose support belongs to $\Omega_{k_2}$. With the notations mentioned above, we have
(i)

$$E_1 = \mathbb{E}_{\mathcal{P}_R} \|g_P - g_{I_1}\|^2 = \frac{1}{|\Omega_{k_1}|} \sum_{I_1 \in \Omega_{k_1}} \sum_{i=1}^{m_0} g_i^2 \mathbb{I}\{i \notin I_1\} = \|g_P\|^2 \left( \frac{1}{\binom{r}{k_1}} \frac{r-k_1}{r} \binom{r}{k_1} \right) = (1 - \frac{k_1}{r})\|g_P\|^2.$$

(ii) Similarly, we have $E_2 = \mathbb{E}_{\mathcal{P}_R} \|g_P - g_{I_2}\|^2 = \frac{1}{|\Omega_{k_2}|} \sum_{I_2 \in \Omega_{k_2}} \sum_{i=1}^{m_0} g_i^2 \mathbb{I}\{i \notin I_2\} = \|g_P\|^2 \left( \frac{1}{\binom{m_0-r}{k_2}} \frac{m_0-r-k_2}{m_0-r} \binom{m_0-r}{k_2} \right) = (1 - \frac{k_2}{m-r})\|g_P\|^2$.

(iii) Denote $E := \mathbb{E}_{\mathcal{P}_R} \|g_P - C_{\mathcal{P}_R,\delta}(g)\|^2$ be the total compression error due to the compressed gradient $C_{\mathcal{P}_R,\delta}(g)$ with respect to $g_P$. Then, by using the linearity of expectation, we have

$$\begin{aligned}
E &= \mathbb{E}_{\mathcal{P}_R} \|g_P - C_{\mathcal{P}_R,\delta}(g)\|^2 \\
&= \mathbb{E}_{\mathcal{P}_R} \|g_P - g_{I_1} \bigoplus g_{I_2}\|^2 \\
&\stackrel{\langle g_P, g_{I_1} \bigoplus g_{I_2}\rangle = \sum_{i \in I_1 \cup I_2} g_i^2}{=} \mathbb{E}_{\mathcal{P}_R} \|g_P\|^2 + \mathbb{E}_{\mathcal{P}_R} \|g_{I_1} \bigoplus g_{I_2}\|^2 - 2\mathbb{E}_{\mathcal{P}_R} \left( \sum_{i \in I_1 \cup I_2} g_i^2 \right) \\
&\stackrel{\langle g_{I_1}, g_{I_2}\rangle = 0}{=} \mathbb{E}_{\mathcal{P}_R} \|g_P\|^2 + \mathbb{E}_{\mathcal{P}_R} \|g_{I_1}\|^2 + \mathbb{E}_{\mathcal{P}_R} \|g_{I_2}\|^2 - 2\mathbb{E}_{\mathcal{P}_R} \left( \sum_{i \in I_1 \cup I_2} g_i^2 \right).
\end{aligned}$$

On the other hand,

$$E_1 + E_2 = 2\mathbb{E}_{\mathcal{P}_R} \|g_P\|^2 + \mathbb{E}_{\mathcal{P}_R} \|g_{I_1}\|^2 + \mathbb{E}_{\mathcal{P}_R} \|g_{I_2}\|^2 - 2\mathbb{E}_{\mathcal{P}_R} \left( \sum_{i \in I_1} g_i^2 \right) - 2\mathbb{E}_{\mathcal{P}_R} \left( \sum_{i \in I_2} g_i^2 \right) = 2\mathbb{E}_{\mathcal{P}_R} \|g_P\|^2 + \mathbb{E}_{\mathcal{P}_R} \|g_{I_1}\|^2 + \mathbb{E}_{\mathcal{P}_R} \|g_{I_2}\|^2 - 2\mathbb{E}_{\mathcal{P}_R} \left( \sum_{i \in I_1 \cup I_2} g_i^2 \right),$$

together with $\mathbb{E}_{\mathcal{P}_R} \|g_P\|^2 \geq 0$ implies $E \leq E_1 + E_2$. □

To measure the compression error due to the compressed gradient $C_{\mathcal{P}_R,\delta}(g)$ with respect to the full gradient vector $g$, by Lemma 8, there exists an $\alpha \in \mathbb{R}$ such that the total expected compression error:

$$\mathbb{E}\|g - C_{\mathcal{P}_R,\delta}(g)\|^2 = E + \sum_{i \in P^c} g_i^2 \leq \alpha \|g\|^2. \tag{7}$$

But there is no guarantee that $\alpha \in [0, 1)$. On the other hand, by Remark 1 we have:

$$\mathbb{E}\|g - C_{\mathcal{P}_R,\delta}(g)\|^2 = \mathbb{E}\|g - \text{Random}r(g)\|^2 \leq (1 - \frac{r}{d})\|g\|^2, \tag{8}$$

which guarantees $C_{\mathcal{P}_R,\delta}$ to be a $\delta$-compressor. Additionally, the sparsifier, $C_{\mathcal{P}_R,\delta}(g)$ is an hybrid sprasifier—It has some attributes of the original sparsifier $C_\delta$, but we are unsure which $k_1$ and $k_2$ random elements are selected via policy $\mathcal{P}_R$. If $k_1 = 0$, then $C_{\mathcal{P}_R,\delta}(g)$ is Random-$r$ [63] sparsifier. If $k_2 = 0$, then $C_{\mathcal{P}_R,\delta}(g)$ is $C_\delta$ sparsifier. Furthermore, if $C_\delta$ is Top-$r$, then $C_{\mathcal{P}_R,\delta}(g)$ is similar to hybrid random-Top-$r$ sparsifier by Elibol et al. [25]. If $C_\delta$ is Top-$r$, $k_1 \leq r, k_2 = 0$, then it is random-Top-$k_1$ sparsifier of Barnes et al. [9].

For inherently sparse vectors $g$, with $C_\delta = I_d$, we have $C_{\mathcal{P}_R,I_d}(g) = g_{I_1}$ and Lemma 8 holds. Moreover, by (7) we have:

$$\mathbb{E}\|g - C_{\mathcal{P}_R,I_d}(g)\|^2 = (1 - \frac{k_1}{r})\|g\|^2. \tag{9}$$

That is, the policy creates a *lossy* compression with compression factor as good as a Random-$k_1$ compressor.

### 10.6 Approximation error due to polynomial fit

**Polynomial regression.** As mentioned in Section 5, over each sorted segment, we apply a polynomial regression. In our experiments, we usually take the degree of the polynomial as 5. The following result concerns piece-wise constant approximation.

**Lemma 9.** [22] (Error of piece-wise constant fit) For $C_S(g) \in C([1, d])$ and $M > 0$ the following are equivalent:
(i) $\text{Var}_{[1,d]}(C_S(g)) \leq M$ and (ii) $\|s - C_S(g)\|_\infty \leq \frac{M}{2p+2}$, for some $s \in S_p^0$— the set of piece-wise constant splines with $p$ knots.

Proof of the above Lemma follows from [22].

**Remark 4.** We can also give a heuristic to calculate $p$ from Lemma 9. First, calculate $M = |(C_S(g)[1] - C_S(g)[2]) - (C_S(g)[d - 1] - C_S(g)[d])|$. By considering the error bound $\frac{M}{2p+2}$ as a function of $p$, we can find closed-form solution for $p$ as $p = \lceil \frac{M}{\sqrt{2}} - 1 \rceil$.

However, for piece-wise linear fit, we propose the following result with an explicit constant.

**Lemma 10.** (Error of piece-wise linear fit with explicit constants) For $C_S(g) \in C^1([1, d])$ with $\text{Var}_{[1,d]}(C_S'(g)) \leq M$, we have $\|s - C_S(g)\|_\infty \leq \frac{2M}{p^2}$, for some $s \in S_p^1$— the set of piece-wise linear splines with $p$ knots.

PROOF. Let $p' = \lceil p/2 \rceil$, the integer part of $p/2$, and let $s^0 \in S_p^0$ satisfy $|C_S'(g)(t) - s^0(t)| \leq \frac{M}{2p'+2}$, as guaranteed by Lemma 9. Denote $V^1 := \int_1^r |C_S'(g)(t) - s^0(t)| dt$. Choose $p'$ knots $\{t_j\}$ in $(1, d)$ such that $\int_{t_j}^{t_{j+1}} |C_S'(g) - s^0(t)| dt = \frac{V^1}{p'+1}$. Define $s(x) = \int_{t_j}^x s^0(t)dt + f(t_j)$ for $x \in [t_j, t_{j+1}]$, $j = 0, 1, ..., p'$. Then $s$ is a piece-wise linear spline function with possible knots at $2p' \leq p$ points such that, for $x \in [t_j, t_{j+1}]$, $|C_S(g)(x) - s(x)| \leq \int_{t_j}^x |C_S'(g)(t) - s^0(t)| dt \leq \int_{t_j}^{t_{j+1}} |C_S'(g)(t) - s^0(t)| dt = \frac{2M}{(2p'+2)^2} \leq \frac{2M}{p^2}$, which implies the result. □

**Remark 5.** Let $s \in S_p^1$. Denote $\hat{\sigma} := \|s - C_S(g)\|$. Then

$$\hat{\sigma} := \|s - C_S(g)\| \leq \sqrt{d}\|s - C_S(g)\|_\infty \overset{\text{By Lemma 10}}{\leq} \frac{2\sqrt{d}M}{p^2}.$$

### 10.7 Compression error from value fitting

Let $\hat{C}(g)$ be the approximation of the sparse vector $C_\delta(g)$ resulted from a $\delta$-compressor. In the intermediate step, we consider the sparse vector $C_S(g)$ that has the components of $C_\delta(g)$ arranged in descending order of magnitude. Let $s \in S_p^1$ be the approximation of $C_S(g)$. Assume no orthogonality and let the angle between $C_\delta(g)$ and $C_S(g)$ be $0 \leq \theta < \pi/2$, and the angle between $s$ and $\hat{C}(g)$ be $0 \leq \theta' < \pi/2$. We aim to calculate the bound on $\mathbb{E}\|\hat{C}(g)\|$, (where $\hat{C}(g)$ is considered to be iteration and worker agnostic) and the Lemma follows. At each node this is what happens:

$$g \to \underbrace{C_\delta(g)}_{\text{Sparse gradient}} \to \underbrace{C_S(g)}_{\text{Rearranged absolute gradient}} \to \underbrace{s}_{\text{Approximated absolute gradient}} \to \underbrace{\hat{C}(g)}_{\text{Reconstructed sparse gradient}}.$$

**Lemma 11.** *(i) If $C_\delta(g)$ is unbiased, that is, $\mathbb{E}(C_\delta(g)) = g$ then*

$$\mathbb{E}\|\hat{C}(g)\|^2 \leq 2(2-\delta)\|g\|^2 + 8(1-\cos\theta)(2-\delta)\|g\|^2 + \frac{32dM^2}{p^4} + \frac{128dM^2}{p^4}(1-\cos\theta') + 32(1-\cos\theta')(2-\delta)\|g\|^2. \tag{10}$$

*(ii) If $C_\delta(g)$ is biased, that is, $\mathbb{E}(C_\delta(g)) \neq g$, then*

$$\mathbb{E}\|\hat{C}(g)\|^2 \leq 2(2-\delta)\|g\|^2 + 16(1-\cos\theta)(2-\delta)\|g\|^2 + \frac{32dM^2}{p^4} + (1-\cos\theta')\frac{128dM^2}{p^4} + 64(1-\cos\theta')(2-\delta)\|g\|^2. \tag{11}$$

Proof. We have

$$\mathbb{E}\|\hat{C}(g)\|^2$$
$$= \mathbb{E}\|\hat{C}(g) - g + g\|^2$$
$$\leq 2\mathbb{E}\|\hat{C}(g) - g\|^2 + 2\|g\|^2.$$

**Case: 1** Consider $C_\delta(g)$ be unbiased, that is, $\mathbb{E}(C_\delta(g)) = g$. Therefore,

$$\mathbb{E}\|g - \hat{C}(g)\|^2$$
$$= \mathbb{E}\|g - C_\delta(g) + C_\delta(g) - \hat{C}(g)\|^2$$
$$\stackrel{\mathbb{E}(C_\delta(g))=g}{=} \mathbb{E}\|g - C_\delta(g)\|^2 + \mathbb{E}\|C_\delta(g) - \hat{C}(g)\|^2$$
$$\stackrel{\text{By (6)}}{\leq} (1-\delta)\|g\|^2 + \mathbb{E}\|C_\delta(g) - \hat{C}(g)\|^2.$$

Further we need to bound $\mathbb{E}\|C_\delta(g) - \hat{C}(g)\|^2$. We have

$$\mathbb{E}\|C_\delta(g) - \hat{C}(g)\|^2$$
$$= \mathbb{E}\|C_\delta(g) - C_S(g) + C_S(g) - \hat{C}(g)\|^2$$
$$\stackrel{\text{By (2)}}{\leq} 2\mathbb{E}\|C_\delta(g) - C_S(g)\|^2 + 2\mathbb{E}\|C_S(g) - \hat{C}(g)\|^2.$$

If the angle between $C_\delta(g)$ and $C_S(g)$ be $0 \leq \theta < \pi/2$, then

$$\mathbb{E}\|C_\delta(g) - C_S(g)\|^2 \stackrel{\text{By Lemma 3}}{=} 2(1-\cos\theta)\mathbb{E}\|C_\delta(g)\|^2$$
$$\stackrel{\text{By Lemma 4}(i)}{\leq} 2(1-\cos\theta)(2-\delta)\|g\|^2. \tag{12}$$

We pause here and quantify: $\|C_S(g) - \hat{C}(g)\|^2$. Let $s \in S_p^1$ be the approximation of $C_S(g)$. Then

$$\mathbb{E}\|C_S(g) - \hat{C}(g)\|^2$$
$$= \mathbb{E}\|C_S(g) - s + s - \hat{C}(g)\|^2$$
$$\stackrel{\text{By (2)}}{\leq} 2\mathbb{E}\|C_S(g) - s\|^2 + 2\mathbb{E}\|s - \hat{C}(g)\|^2$$
$$\stackrel{\text{By Remark 5}}{\leq} \frac{8dM^2}{p^4} + 2\mathbb{E}\|s - \hat{C}(g)\|^2 \tag{13}$$

Similar to before, if the angle between $s$ and $\hat{C}(g)$ be $0 \leq \theta' < \pi/2$, then

$$\mathbb{E}\|s - \hat{C}(g)\|^2 \stackrel{\text{By Lemma 3}}{=} 2(1-\cos\theta')\mathbb{E}\|s\|^2$$
$$\leq 2(1-\cos\theta')\mathbb{E}\|s - C_S(g) + C_S(g)\|^2$$
$$\stackrel{\text{By (2)}}{\leq} 4(1-\cos\theta')\|s - C_S(g)\|^2 + 4(1-\cos\theta')\mathbb{E}\|C_S(g)\|^2$$
$$\stackrel{\mathbb{E}\|C_\delta(g)\|^2=\mathbb{E}\|C_S(g)\|^2}{=} (1-\cos\theta')\frac{16dM^2}{p^4} + 4(1-\cos\theta')\mathbb{E}\|C_\delta(g)\|^2$$
$$\stackrel{\text{By Lemma 4}(i)}{\leq} (1-\cos\theta')\frac{16dM^2}{p^4} + 4(1-\cos\theta')(2-\delta)\|g\|^2. \tag{14}$$

Therefore,

$$\mathbb{E}\|C_\delta(g) - \hat{C}(g)\|^2$$
$$\leq 2\mathbb{E}\|C_\delta(g) - C_S(g)\|^2 + 2\mathbb{E}\|C_S(g) - \hat{C}(g)\|^2$$
$$\leq 4(1-\cos\theta)(2-\delta)\|g\|^2 + \frac{16dM^2}{p^4} + \frac{64dM^2}{p^4}(1-\cos\theta') + 16(1-\cos\theta')(2-\delta)\|g\|^2$$

Combining all together we have

$$\mathbb{E}\|\hat{C}(g)\|^2$$
$$\leq 2(2-\delta)\|g\|^2 + 8(1-\cos\theta)(2-\delta)\|g\|^2 + \frac{32dM^2}{p^4} + \frac{128dM^2}{p^4}(1-\cos\theta') + 32(1-\cos\theta')(2-\delta)\|g\|^2.$$

**Case: 2** If $C_\delta(g)$ be biased, that is, $\mathbb{E}(C_\delta(g)) \neq g$ then

$$\mathbb{E}\|g - \hat{C}(g)\|^2$$
$$\leq 2\mathbb{E}\|g - C_\delta(g)\|^2 + 2\mathbb{E}\|C_\delta(g) - \hat{C}(g)\|^2$$
$$\stackrel{\text{By (6)}}{\leq} 2(1-\delta)\|g\|^2 + 2\mathbb{E}\|C_\delta(g) - \hat{C}(g)\|^2.$$

For biased compressor $C_\delta(g)$, we have From Lemma 4 (ii)

$$\mathbb{E}\|C_\delta(g)\|^2 \leq 2(2-\delta)\|g\|^2.$$

As a result,

$$\mathbb{E}\|C_\delta(g) - C_S(g)\|^2 \stackrel{\text{By Lemma 3}}{=} 2(1-\cos\theta)\mathbb{E}\|C_\delta(g)\|^2$$
$$\stackrel{\text{By Lemma 4}(ii)}{\leq} 4(1-\cos\theta)(2-\delta)\|g\|^2. \tag{15}$$

and

$$\mathbb{E}\|s - \hat{C}(g)\|^2 \stackrel{\text{By Lemma 3}}{=} 4(1-\cos\theta')\|s - C_S(g)\|^2 + 4(1-\cos\theta')\mathbb{E}\|C_\delta(g)\|^2$$
$$\stackrel{\text{By Lemma 4}(i)}{\leq} (1-\cos\theta')\frac{16dM^2}{p^4} + 8(1-\cos\theta')(2-\delta)\|g\|^2. \tag{16}$$

Finally,

$$\mathbb{E}\|C_\delta(g) - \hat{C}(g)\|^2$$
$$\leq 2\mathbb{E}\|C_\delta(g) - C_S(g)\|^2 + 2\mathbb{E}\|C_S(g) - \hat{C}(g)\|^2$$
$$\leq 8(1-\cos\theta)(2-\delta)\|g\|^2 + \frac{16dM^2}{p^4} + (1-\cos\theta')\frac{64dM^2}{p^4} + 32(1-\cos\theta')(2-\delta)\|g\|^2 \tag{17}$$

Combining all together we have

$$\mathbb{E}\|\hat{C}(g)\|^2$$
$$\leq 2\mathbb{E}\|\hat{C}(g) - g\|^2 + 2\|g\|^2$$
$$\leq 2(2-\delta)\|g\|^2 + 16(1-\cos\theta)(2-\delta)\|g\|^2 + \frac{32dM^2}{p^4} + (1-\cos\theta')\frac{128dM^2}{p^4} + 64(1-\cos\theta')(2-\delta)\|g\|^2.$$

Hence the result. □

## 10.8 Comment on convergence.

Lemma 11 shows that the fitted sparse gradients are bounded and is instrumental for proving the convergence of distributed SGD. One can find the bound on compressed aggregated gradient $\tilde{g}_k$ resulting at $k^{\text{th}}$ iteration. Denote $\hat{C}(g_k^i)$ be the approximation of the sparse vector $C_\delta(g_k^i)$ resulted from a $\delta$-compressor at $i^{\text{th}}$ worker, at the $k^{\text{th}}$ iteration. Denote the compressed aggregated gradient at $k^{\text{th}}$ iteration to be $\tilde{g}_k := \frac{1}{n}\sum_{i=1}^n \hat{C}(g_k^i)$. By using standard technique we can show that $\mathbb{E}\|\tilde{g}_k\|^2$ is also bounded for both biased and unbiased $C_\delta(g)$.

With this, based on the strong growth condition of stochastic gradients [24, 69], for a lower bounded, Lipschitz smooth, and non-convex loss function $f$, following [24], the distributed SGD with an $\delta$ sparsifier where the sparse values are approximated by a linear regressor converges, that is, $\min_{t\in[T]} \mathbb{E}(\|\nabla f_t\|^2) \to 0$ as $T \to \infty$.

## 11 Additional experimental results

*11.0.1 Implemented methods.* We implement DeepReduce[2] as an extension of GRACE [75], a framework that supports many popular sparsification techniques and interfaces with various low-level communication libraries for distributed deep learning. Table 3 presents a summary of the methods we implement.

**Run Length Encoding (RLE).** Since RLE is a lossless method designed for continuous repetitive symbols, it is not directly applicable to non-repetitive gradient indices. We convert gradient indices into bitmap format, which is a boolean bit string indicating which elements are selected. In this way, RLE can be used to encode the continuous zeros and ones in the bitmap. Note that, the compression rate is highly dependent on the distribution of the indices. That is, RLE is more beneficial if gradient indices contain more continuous integers.

---
[2]Available at: https://github.com/hangxu0304/DeepReduce

**Table 3: Summary of implementations.** $\mathrm{DR}_{\mathrm{idx}}^{\mathrm{val}}$ denotes instantiation of DeepReduce with *idx* and *val* as index and value compression method, respectively.

| Method | Idx | Val | Device | Framework |
|---|---|---|---|---|
| $\mathrm{DR}_{\mathrm{BF-Naive}}^{\varnothing}$ | ✓ | | CPU | TFlow |
| $\mathrm{DR}_{\mathrm{BF-P0}}^{\varnothing}$, $\mathrm{DR}_{\mathrm{BF-P1}}^{\varnothing}$ | ✓ | | GPU | PyTorch |
| $\mathrm{DR}_{\mathrm{BF-P0}}^{\varnothing}$, $\mathrm{DR}_{\mathrm{BF-P1}}^{\varnothing}$, $\mathrm{DR}_{\mathrm{BF-P2}}^{\varnothing}$ | ✓ | | CPU | TFlow, PyTorch |
| $\mathrm{DR}_{\mathrm{RLE}}^{\varnothing}$ | ✓ | | CPU | TFlow, PyTorch |
| $\mathrm{DR}_{\mathrm{Huffman}}^{\varnothing}$ | ✓ | | CPU | PyTorch |
| $\mathrm{DR}_{\varnothing}^{\mathrm{Fit-Poly}}$ | | ✓ | GPU, CPU | TFlow, PyTorch |
| $\mathrm{DR}_{\varnothing}^{\mathrm{Fit-DExp}}$ | | ✓ | GPU | TFlow |
| $\mathrm{DR}_{\varnothing}^{\mathrm{Deflate}}$ | | ✓ | CPU | PyTorch |
| $\mathrm{DR}_{\varnothing}^{\mathrm{QSGD}}$ | | ✓ | GPU | PyTorch |
| $\mathrm{DR}_{\mathrm{BF-P0}}^{\mathrm{Fit-Poly}}$, $\mathrm{DR}_{\mathrm{BF-P1}}^{\mathrm{Fit-Poly}}$ | ✓ | ✓ | GPU | PyTorch |
| $\mathrm{DR}_{\mathrm{BF-P2}}^{\mathrm{QSGD}}$ | ✓ | ✓ | GPU | PyTorch |
| 3LC | | ✓ | GPU | TFlow |
| SketchML | | ✓ | CPU | PyTorch |
| SKCompress | ✓ | ✓ | CPU | PyTorch |

**Huffman Encoding.** The key idea of Huffman Coding is to use fewer bits to represent more frequent symbols. We note that most indices are much smaller than $2^{32}$, and consequently their binary format start with continuous zero bits. Based on this observation, we can use Huffman Coding to compress the binary format of each index to remove the redundancy. The codec is constructed from all possible indices of the target model. (i.e. If the targets gradient size is $d$, then we use $0 \sim d-1$ for codec construction). The encoding phase contains 2 steps: unpack each 32-bit integer gradient key into Byte format and then encode each index with the pre-defined codec. The decoding phase is just a reversed process.

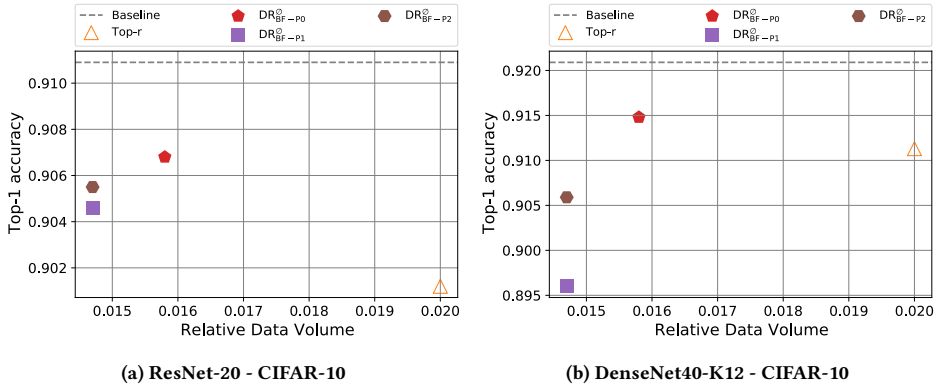

(a) ResNet-20 - CIFAR-10  (b) DenseNet40-K12 - CIFAR-10

**Figure 15:** (a) Data volume vs. accuracy of ResNet-20 on CIFAR-10. (b) Data volume vs. accuracy of DensNet40 on CIFAR-10. We compare $\mathrm{DR}_{\mathrm{BF}}^{\varnothing}$ (with P0, P1, P2, naive policy) against Top-$r$ and baseline. Ratios of Top-$r$ are 1% for ResNet20, and 0.5% for DenseNet40. FPR is set to 0.001.